\documentclass[english]{IEEEtran}
\usepackage[T1]{fontenc}
\usepackage[latin9]{inputenc}
\usepackage{color}
\usepackage{url}
\usepackage{amsmath}
\usepackage{graphicx}
\PassOptionsToPackage{normalem}{ulem}
\usepackage{ulem}

\makeatletter
\numberwithin{figure}{section}

\usepackage{cite}
\counterwithout{figure}{section}
\counterwithout{table}{section}

\makeatother

\usepackage{babel}
\begin{document}
\title{Investigating classification learning curves for automatically generated
and labelled plant images}

\author{\IEEEauthorblockN{Michael A. Beck\IEEEauthorrefmark{1}, Christopher P. Bidinosti, Christopher J. Henry, Manisha
Ajmani\\}
\IEEEauthorblockA{University of Winnipeg, Winnipeg, MB Canada\\
Email: \IEEEauthorrefmark{1}m.beck@uwinnipeg.ca}}
\maketitle
\begin{abstract}
In the context of supervised machine learning a learning curve describes
how a model's performance on unseen data relates to the number of
samples used to train the model. In this paper we present a dataset
of plant images with representatives of crops and weeds common to
the Manitoba prairies at different growth stages. We determine the
learning curve for a classification task on this data with the ResNet
architecture. Our results are in accordance with previous studies
and add to the evidence that learning curves are governed by power-law
relationships over large scales, applications, and models. We further
investigate how label noise and the reduction of trainable parameters
impacts the learning curve on this dataset. Both effects lead to the
model requiring disproportionally larger training sets to achieve
the same classification performance as observed without these effects.
\end{abstract}

\section{Introduction\label{sec:Introduction}}

Image classification through convolutional neural networks (CNN) became
a staple of today's machine learning discussion. Here, the utilization
of GPUs as well as the availability of large, open-access datasets
enabled the explosive success of CNNs. Some examples of these datasets
are MNIST \cite{lecun1998mnist,deng2012mnist}, CIFAR \cite{krizhevsky2009learning},
and ImageNet \cite{ILSVRC15}. Fuelled by computational power, improved
architectures, and data ML has made considerable progress. However
the need for data persists up to today. Indeed, there is a general
``more is better'' mentality when it comes to the amount of training
samples available. Yet, at the same time data-generation, labelling,
storage, dissemination, and usage comes with non-neglectable costs
in time, infrastructure, and money (see for example \cite{najafabadi2015deep}).
Consequently, it becomes increasingly important to answer the following
questions: 
\begin{itemize}
\item How much data is needed to achieve a certain performance goal?
\item How does performance relate to sample size?
\item Can available data be reduced to a subset without impairing the performance
of the models trained on it? Can the removal of samples even improve
model performances?
\item How can such a reduction of data be performed? How can we valuate
data points within a training set and determine which data points
can or should be discarded?
\end{itemize}
\begin{figure}
\begin{center}\includegraphics[width=1\columnwidth]{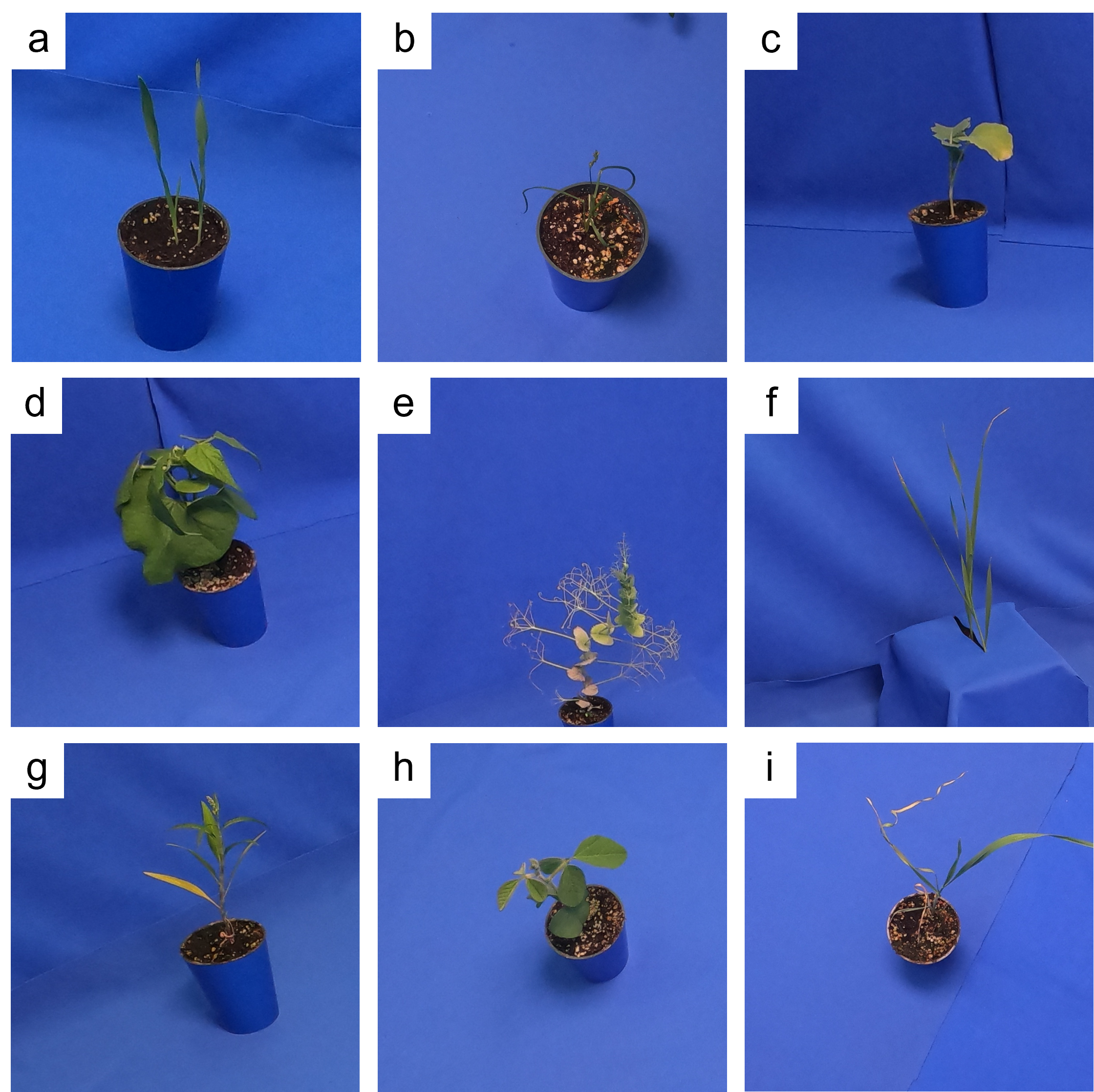}\end{center}\caption{Examples images of the dataset considered in this paper: (a) Barley,
(b) Barnyard Grass, (c) Canola, (d) Drybean, (e) Field Pea, (f) Oat,
(g) Smartweed, (h) Soybean, (i) Wheat\label{fig:Examples-images}}
\end{figure}
\begin{figure*}[t]
\begin{center}\includegraphics[width=0.5\textwidth]{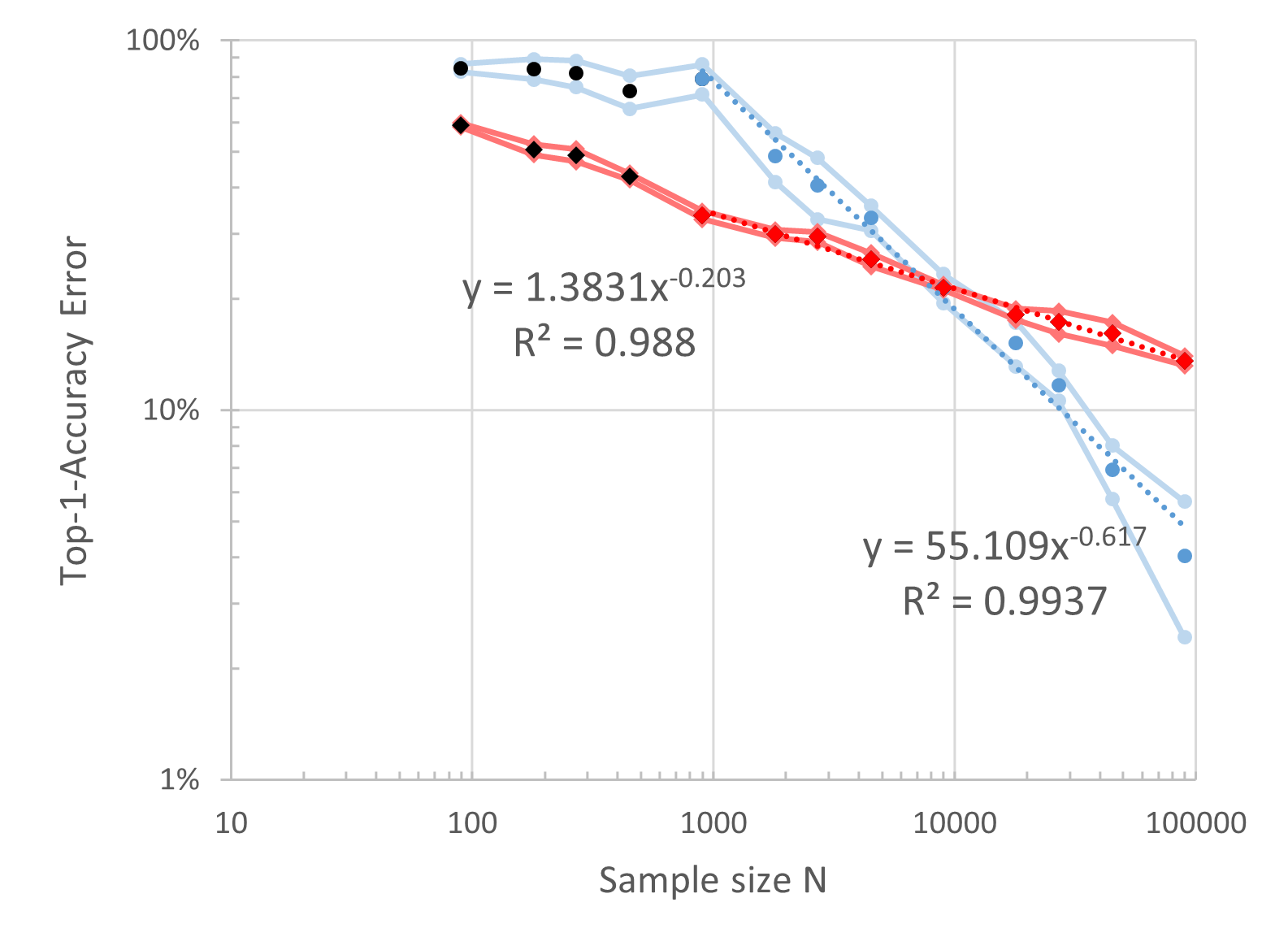}\includegraphics[width=0.5\textwidth]{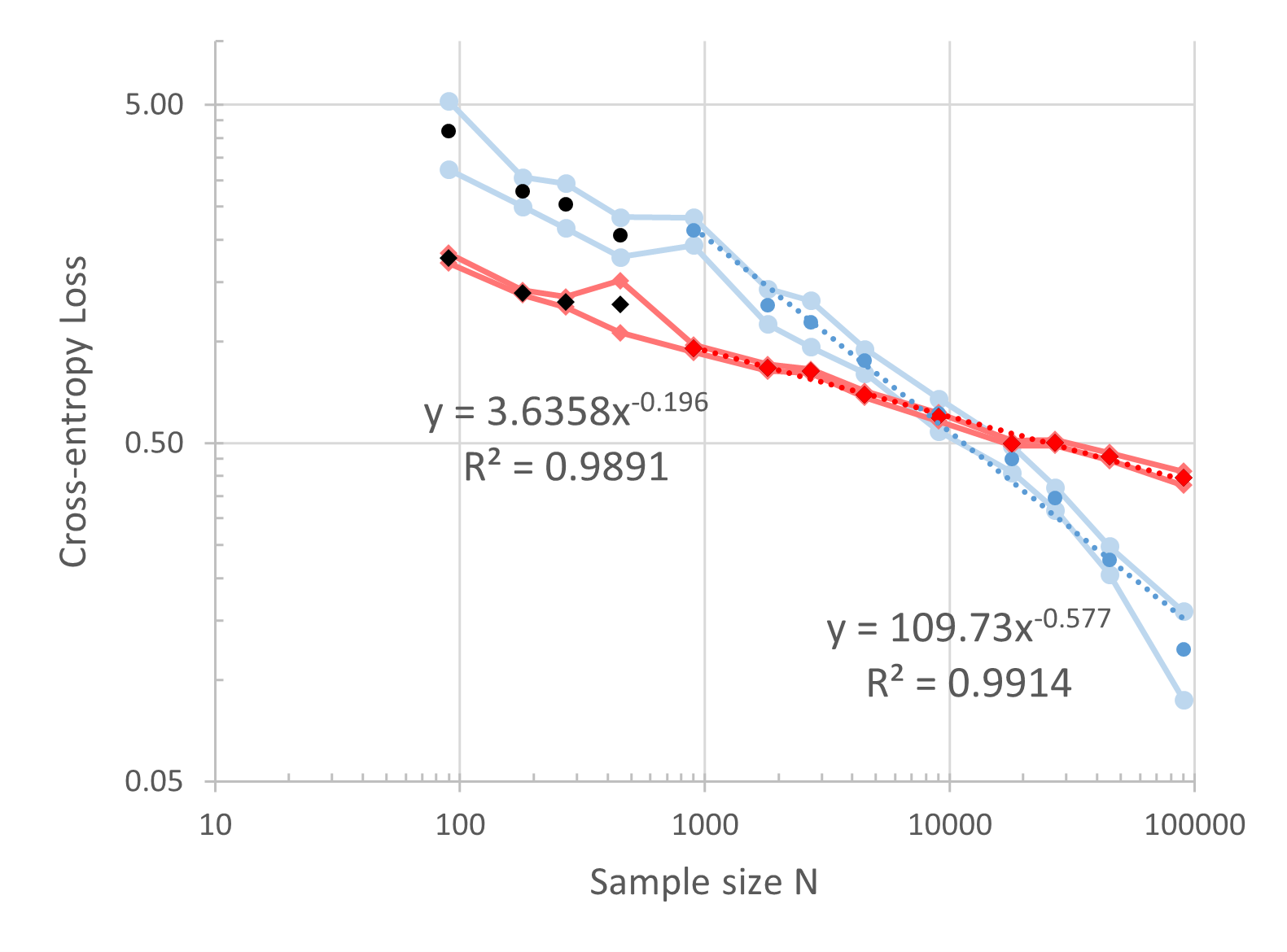}\end{center}\caption{Learning curves for a ResNet50 architecture trained on our dataset for top-1 accuracy error (left) and cross-entropy loss (right). Circular markers correspond to a ResNet50 models with random initializiation. Diamond-shaped markers correspond to a ResNet50 model pretrained on ImageNet. The light colored curves represent one standard deviation around the mean, respectively. We identify the power-law behavior to start at a sample size of $N=900$, indicated by the color change of the curves. The power-law fits are visualized as dotted curves. \label{fig:random}} \end{figure*}

Due to the black-box nature of modern deep neural networks and the
high-dimensionality of images the above questions are difficult to
address directly. It is thus useful to make large-scale observations,
one of these is the concept of learning curves. A learning curve plots
a model's performance on a held out testing set (this metric is often
called risk)\emph{ }against the number of samples the model was trained
on. Several studies have observed a power-law relationship between
risk and training volume \cite{hestness2017deep,rosenfeld2019constructive,spigler2020asymptotic,bahri2021explaining}
with exponents usually in the range of $\alpha\in(-1,0)$. There are
theoretical discussions that suggest that the values of $\alpha$
depends on the inherent dimensionality of the sample data \cite{bahri2021explaining}.
It must also be said that the power-law relationship does not hold
on all sample scales. Instead we can observe three phases for learning
curves \cite{hestness2017deep}, which are in order of increasing
size of training sets: (i) the small data phase, in which the model
does not perform significantly better than making random predictions;
(ii) the phase in which the power-law relationship holds; (iii) the
phase of irreducible error, in which the power-law relation comes
to an end and no further improvement can be observed. Even though
the learning curve transitions into and out of the power-law behavior
this description makes model performances somewhat predictable. The
basic idea is this: We can train the model on a sequence of relatively
small subsets and determine the exponent $\alpha$. Then we can use
the power-law relation to estimate the model's performance for the
full dataset (assuming we do not transition into the phase of irreducible
error in the meantime!) \cite{rosenfeld2019constructive}. This idea
can be extended to more elaborate methods. See \cite{viering2021shape}
for an overview. 

In this paper we investigate a dataset of plant images on blue background,
see Figure \ref{fig:Examples-images} for an example. The sample consists
of 90900 images: 10000 training images and 100 validation images per
9 different plant classes. The images show a variety of different
growth stages for each class and cover a wide range of imagine angles.
The images are randomly selected from a larger collection that was
created and labelled by an autonomous robotic imager \cite{10.1371/journal.pone.0243923}.
The machine learning task we consider is to correctly classify the
images to one of the 9 plant species. In this context we perform the
following analysis: 
\begin{itemize}
\item We confirm the power-law relationship between training volume and
model accuracies and measure the respective exponent $\alpha$. 
\item We investigate how the introduction of noise (purposefully mislabelling
a random subset of training samples) affects $\alpha$.
\item We re-evaluate $\alpha$ by changing the model from a randomly initialized
one to a model pretrained on ImageNet. This reduces the number of
trainable parameters by several orders of magnitude. 
\end{itemize}
These are necessary first steps to describe the data-quality with
respect to training effectiveness. Our observations indicate that
noise has a significant impact on the exponent $\alpha$. Furthermore,
we can observe that a reduction of training parameters leads to a disadvantegeous $\alpha$, which in turn leads to reduced model performance on large training volume. 

\section{Description of Dataset\label{sec:dataset}}

\begin{figure*}[t]
\begin{center}\includegraphics[width=0.5\textwidth]{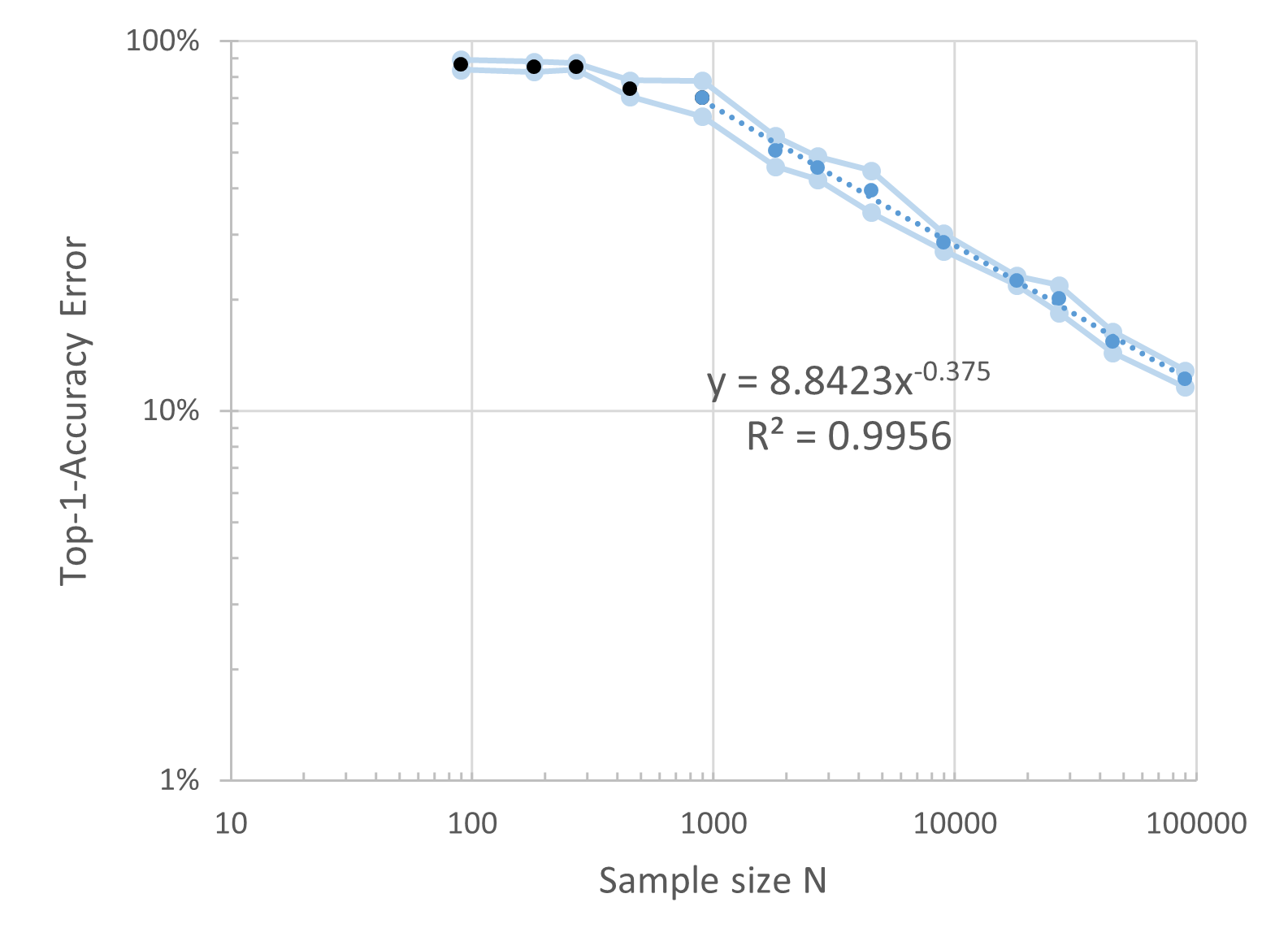}\includegraphics[width=0.5\textwidth]{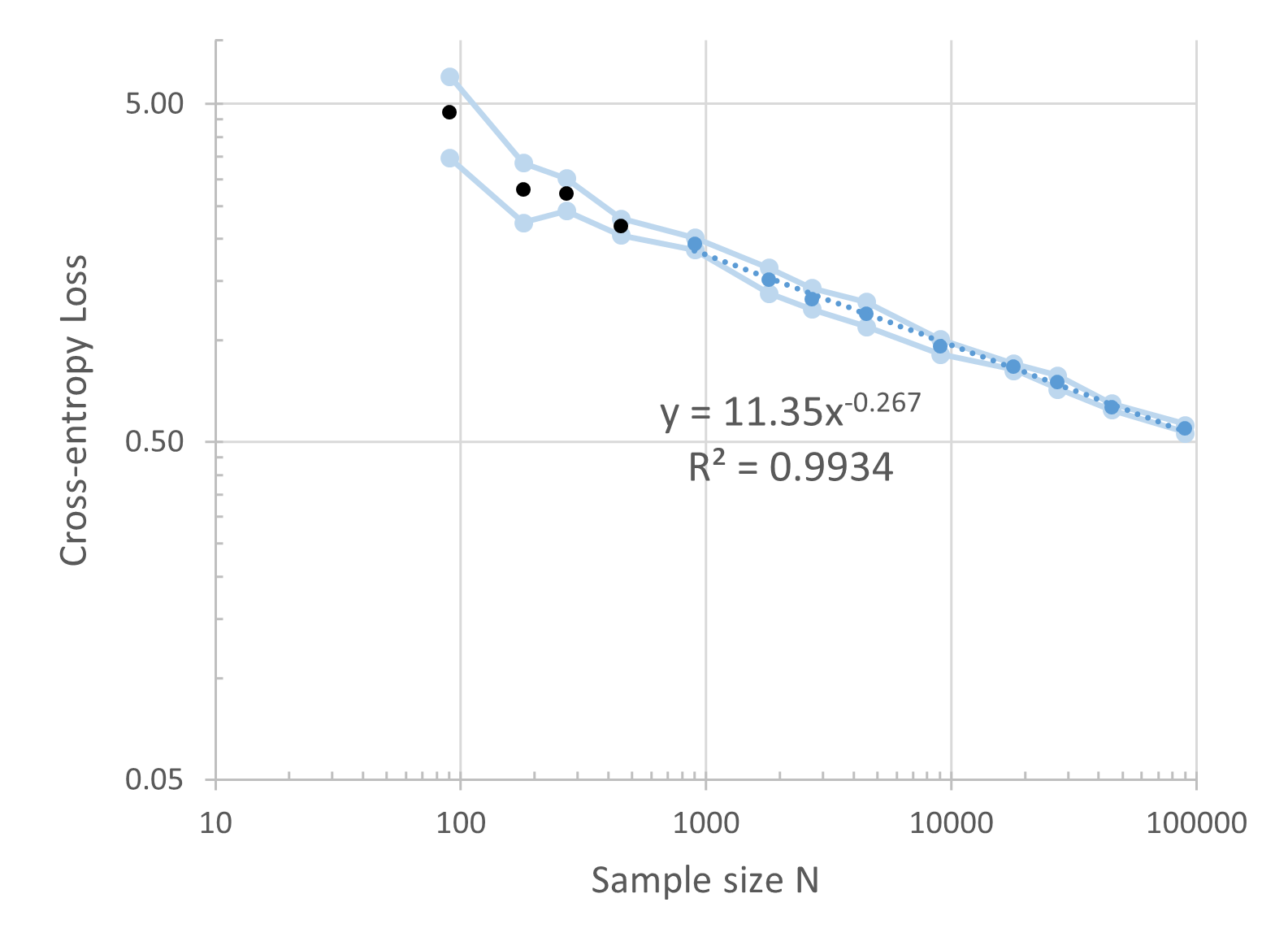}\end{center}\caption{Learning curves after introducing 5\% label noise to the sample data
for top-1 accuracy error (left) and cross-entropy loss (right). The
overall performance of the model declines, whereas the power-law exponent
becomes shallower compared to models trained on the original data.
\label{fig:noise}}
\end{figure*}

In this paper we consider plant images taken by our robotic imager
EAGL-I (see \cite{10.1371/journal.pone.0243923}). This system is
capable of automatically imaging and labelling several plants at once
from a wide variety of angles and distances. Examples of such images
can be seen in Figure \ref{fig:Examples-images}. The overall purpose
of these images is to support machine learning applications in digital agriculture,
such as weed-detection, yield-estimates, plant-health, and phenotyping.
The whole dataset (more than 1.5 million images of the type seen in
Figure \ref{fig:Examples-images}, plus over half a million additional
unlabelled field-images and over half a million labelled images that
contain multiple plants) is available via the TerraByte project homepage\footnote{\textcolor{blue}{\uline{\url{https://terrabyte.acs.uwinnipeg.ca/resources.html}}}.
See also \cite{beck2022terrabyte} for more details on the downloader.}. More data as well as data-types are being continuously added. 

Out of this whole dataset we have randomly selected 90900 images such
that there are 10100 images per plant class. The classes we chose
contain 7 different cash-crops (e.g., barley, wheat, soybeans, peas)
and 2 types of weeds (barnyard grass and smartweed) that are common
to the Manitoba region (see also Figure \ref{fig:Examples-images}. Thus, the dataset consists of 90000 sample
images and 900 test images. Throughout this paper the 900 test images
are the same for all risk evaluations and have not been used in the
training of any of the models. The 90000 sample images are shuffled
per class. This ensures that two images of the same individual plant
imaged on the same day are not listed in order when creating the training
subsets. Then we created an increasing sequence of subsets, such that
each larger subset contains all the images of the previous subsets.
The smallest sample subset contains 10 images per class (90 images
in total) whereas the largest subset is the entire sample set of 10000
images per class. We will denote these subsets by $T_{N}$, where
$N$ is the size of the subset, for example $T_{900}$ or $T_{45K}=T_{45000}$.

\section{Training Setup\label{sec:Training} }

We note that results can depend on model-architecture, model-size,
and tuning of hyper-parameters. However, at this point we are foremost
interested in the data-quality, a description of data-redundancy,
and learning curve parameters. The literature suggests that we can
expect not identical, yet very similar, results when choosing a different
model to train on the same data (see for example \cite{rosenfeld2019constructive}).
Unless mentioned otherwise we always trained a classification model
as follows: 
\begin{itemize}
\item We use the popular \emph{ResNet50} architecture \cite{he2016deep}
for image classification with ADAM optimizer.
\item We choose cross-entropy or top-1 accuracy as loss function. 
\item The model's weights have been chosen randomly. As the classes are
balanced, there is no need for class weights. 
\item 20\% of the training set $T_{N}$ is held out as validation data.
It is used to determine when the model has converged. In models with
equal training volume the same data is held out.
\item We use early stopping on the validation-accuracy with a patience of
50 epochs, and a maximum of 100 training epochs.
\item The only data-augmentation used is a 50\% chance to horizontally flip
the image and rescaling to 224x224 pixels.
\item Validation accuracy and loss is either reported for the model that
triggered the early stopping, or the best model found. 
\item To evaluate the trained models, we classify 900 images of the overall
held out test set. This is independent of $N$, the volume of the
training set and validation set used.
\item Due to random choices each configuration of training is repeated 5 times. We plot the mean loss of these 5 models, as well as plus/minus one standard deviation from the mean.
\end{itemize}

\section{Results\label{sec:Results} }

\subsection{Learning curve exponent}

Recent research \cite{hestness2017deep,rosenfeld2019constructive,spigler2020asymptotic,bahri2021explaining}
observed a power-law behavior for learning curves of the form: 
\[
L(N)=c\cdot N^{\alpha}
\]
with $\alpha\in(-1,0)$. For example \cite{hestness2017deep} reports
a value of $\alpha=-0.35$ after training a family of ResNet-models
on ImageNet data. We note here that power-law relationships appear
as a linear curve in log-log-plots with slope $\alpha$, due to 
\begin{align}
	\log(L(N))=\log(c)+\alpha\cdot\log(N).\label{eq:log-linear}
\end{align}
To estimate the power-law exponent, we follow the procedure, which we assume is also followed by 
the authors of previous relevant work (e.g., \cite{hestness2017deep}). This method uses a least-square 
fit of a linear model  in the log-log space, i.e. after applying Equation (\ref{eq:log-linear}). 
That said, we also fitted a non-linear model of the form \(L(N)=c\cdot N^{\alpha}\) directly to the data, 
minimizing least-squares and found discrepancies up to 8\%. For a discussion and techniques on estimating exponents of pawer-laws we refer to \cite{hanel2017exponentML}.

To validate that a power-law can be observed in our dataset,
we train our ResNet-model on several sample sets $T_{N}$ . The result
is illustrated in Figure \ref{fig:random} it shows the model's performance
over the full range of training subsets on a log-log-plot. From the
observed accuracies and cross-entropy loss we identify the power-law
region starting not earlier than at the subset of size $N=900$. Further
we did not observe reaching the phase of irreducible error. We fitted
a power-law curve over the datapoints from $N=900$ to $N=90000$
and observe an exponent of $\alpha\approx-0.62$ for the top-1 accuracy
error and $\alpha\approx-0.58$ for the cross-entropy loss. These values
are relatively large -- at least compared to $\alpha=-0.35$ for ResNet
models on ImageNet \cite{hestness2017deep}. Having a large exponent
tells us that training a classification task on our data is efficient, 
whereas the reasons for that can be manifold. We suspect three
data characteristics working together towards this effect: (i) For
once, the representatives of the classes in our dataset have likely
less variance than the classes in ImageNet. For example, two different
soybeans differ mostly in the number and position of their leaves
and even those correlate strongly with the plants maturity. For comparison
a typical class in ImageNet, say ``Bicycle'', can be represented
by a wide range of colors and forms. (ii) The objects to classify
in the plant dataset are all in front of a homogenous, mostly noise-free,
unicolored background. This is not the case in ImageNet where the
objects to classify are placed in natural environments. (iii) Our
images had been automatically labelled and thus have virtually no labelling
errors. ImageNet, instead, was manually labelled and for early versions
of ImageNet (on which the exponent in \cite{hestness2017deep} is
based) it is estimated that 5\% of the images are mislabelled \cite{northcutt2021pervasive}.
We investigate now in more detail the impact of noisy labels for our
data.

\subsection{Noisy training data}

In this evaluation we purposefully change the labels in $T_{N}$,
before using them in training. To be more specific: In $T_{90K}$
we changed for each datum with a 5\% chance the label to a random
label selected from one of the 8 possible wrong labels. Consequently,
if a datum was changed in $T_{90K}$ it is changed in the same way
for each subset $T_{N}$ it is a member of. Since the validation set
is a subset of $T_{N}$ the introduced noise spreads over both the
training and the validation set. We did not introduce noise to the
held out test set of 900 images. Figure \ref{fig:noise} shows the
learning curves for top-1 accuracy error and cross-entropy loss, respectively.
As expected, the overall performance on the test set has suffered,
compared to using the original training data. More noteworthy, however,
is that introducing 5\% noise has resulted in a significant increase
of the exponent $\alpha$ and thus a reduced effectiveness of the
training data. For the top-1 accuracy error the difference is 0.242,
for the cross-entropy loss the difference is 0.31. Thus, noisy training
data has affected the effectiveness of the training data on all scales
and we need disproportionally more data to compensate this effect
and achieve comparable performance. 

\subsection{Reduction of model parameters}

To investigate how the exponent $\alpha$ changes when modifying the
amount of training parameters, we perform a simple change in the ResNet
model. Instead of starting the model from scratch with randomly initialized
weights we load a model that has already been trained on ImageNet
and freeze all layers, but the output layer. This is a commonly used method to avoid overfitting, as it reduces the number
of trainable parameters. In our case the number of parameters is reduced from 23.5 million  to 18 thousand. 
The effect on the learning curve is illustrated in Figure
\ref{fig:pretrained}. We can observe two effects on the models' performances: First, the performance on small training volumes is increased, when reducing the trainable parameter space.
However, the two learning curves intersect roughly at \(T_{9K}\) and the original training method (using an randomly initialized model) achieves better performance. 
Keeping in mind that learning curves seem to be predictable by fitting power laws this intersection point can be predicted by training both models on small data volumes.
Or stated differently: If we are only interested in selecting the best of several models for a large dataset, we can train these on a short sequence increasingly larger, yet still small datasets.
This serves
as an example of how learning curves can lead to significant time savings for model selection.

\section{Conclusion\label{sec:Conclusion-and-Future}}

We investigated learning curves for a dataset that consists of different
crops and weeds common to the Manitoba prairies. We first observed
that the learning curves for our data follows a power-law relation
with a small exponent $\alpha\approx-0.58$ (cross-entropy loss), indicating
that the classification task is comparably easy to other public datasets,
such as ImageNet. We then investigated how the introduction of labelling
noise or a reduction of trainable parameters influences the exponent.
Both resulted in a significant increase of the exponent and thus a
disproportionally larger amount of data is required to achieve results
comparable to the first scenario (no noise, randomly initialized weights).
By comparing the parameters of learning curves for different models
on the same dataset one can quickly determine which models are more
suitable for the task at hand. We invite researchers to analyze our
dataset further. It is available under: \textcolor{blue}{\uline{\url{https://terrabyte.acs.uwinnipeg.ca/resources.html}}}

\bibliographystyle{unsrt}
\bibliography{refs/references}

\end{document}